\documentclass[conference]{IEEEtran}
\usepackage{cite}
\usepackage{algorithmic}
\usepackage{amsmath,graphicx,hyperref}
\usepackage{graphicx}    
\usepackage{caption}     
\usepackage{subcaption}  
\usepackage{url}
\usepackage{tabularx}
\usepackage{xcolor}
\usepackage{textcomp}
\usepackage[utf8]{inputenc} 
\usepackage[T1]{fontenc}    
\usepackage{amsfonts}       
\usepackage{nicefrac}       
\usepackage{microtype}      
\usepackage{textcomp}
\usepackage{float}
\def\BibTeX{{\rm B\kern-.05em{\sc i\kern-.025em b}\kern-.08em
    T\kern-.1667em\lower.7ex\hbox{E}\kern-.125emX}}

\graphicspath{{Figure/}}

\begin{document}
%
\title{TRIPROBE: Probing Task Separability Beyond Classification for XAI}

\author{\IEEEauthorblockN{1\textsuperscript{st} Amirhossein Sadough}
\IEEEauthorblockA{\textit{Machine Learning and Neural Computing} \\
\textit{ Radboud University}\\
Nijmegen, Netherlands\\
ORCID:0009-0005-5647-2888}
\and
\IEEEauthorblockN{2\textsuperscript{nd} Freek Hens}
\IEEEauthorblockA{\textit{Machien Learning and Neural Computing} \\
\textit{Radboud University}\\
Nijmegen, Netherlands\\
ORCID:0009-0005-0405-0560 }
\and
\IEEEauthorblockN{3\textsuperscript{rd} Aleksa Bokšan}
\IEEEauthorblockA{\textit{Delft University of Technology} \\
Delft, Netherlands\\
ORCID:0009-0003-3099-4472}
\and
\IEEEauthorblockN{4\textsuperscript{th} Mohammad Mahdi Dehshibi}
\IEEEauthorblockA{\textit{Unconventional Computing Lab} \\
\textit{University of the West of England (UWE)}\\
Bristol, United Kingdom \\
ORCID: 0000-0001-8112-5419}
\and
\IEEEauthorblockN{5\textsuperscript{th} Mahyar Shahsavari}
\IEEEauthorblockA{\textit{Machine Learning and Neural Computing} \\
\textit{Radboud University}\\
Nijmegen, Netherlands \\
ORCID:0000-0002-9671-0917}

}
\maketitle
\begin{abstract}
Modern evaluation of learning pipelines often reduces to downstream accuracy, leaving open the question of why tasks succeed or fail.
TriProbe addresses this gap with a multi-level probing framework for explainable diagnosis of task separability. Rather than treating models as black boxes, TriProbe traces how separability evolves across inputs, learned features, and final classifiers. It decomposes multi-task problems into binary subtasks and applies three complementary probes: a Foundational Probe on input spaces, a Latent Probe on feature representations, and a Final Probe on classifier outputs. Using Maximum Fisher’s Discriminant Ratio as a principled separability metric, TriProbe identifies bottlenecks and affected task pairs. Experiments on the Roshambo sEMG benchmark show how TriProbe reveals hidden breakdowns, guiding data collection, validation, and architecture design.
\end{abstract}

\section{Introduction}
Understanding why learning models succeed or fail in distinguishing between tasks remains a fundamental challenge in machine learning and signal processing. Across diverse application areas, including speech, EEG/MEG, and sEMG, task separability is affected not only by intrinsic data complexity but also by architectural choices across the learning pipeline \cite{r2026}. Existing evaluation methods typically focus on downstream accuracy, providing little insight into \emph{where} separability bottlenecks emerge \cite{widm-1493,accuracy-explainability2022,exAI-2021}. This lack of interpretability complicates both data collection and model design, motivating the need for systematic and explainable tools \cite{goh_artificial_2021,AIEX_review23}.  

Model probing has emerged as a promising methodology for understanding how representations evolve in modern learning pipelines. Early work introduced diagnostic classifiers to test whether embeddings captured linguistic or structural information \cite{alain2016linear, hewitt2019structural,ref_P}. Complementary explainable AI approaches, such as Shapley values and integrated gradients, provide attribution scores but typically stop short of tracing separability across multiple stages of the pipeline \cite{sundararajan2020many, sundararajan2017axiomatic}. In parallel, studies in physiological signals, such as sEMG gesture recognition, have emphasized the importance of transfer learning and representation quality for task separability \cite{ref_D,YADAV2025103207}. However, these methods primarily aim at boosting accuracy rather than systematically diagnosing \emph{where} separability is gained or lost across stages.  

To address this gap, we introduce \textbf{TriProbe}, a multi-level probing framework that explains the root causes of task separability difficulties by tracing them across successive stages of learning. 
The central premise of TriProbe is that class separability may either deteriorate or improve throughout different stages of the learning pipeline. Understanding these dynamics requires not only identifying where representational bottlenecks arise but also determining which task pairs are most affected. 

{TriProbe} decomposes a multi-task classification problem into a set of binary sub-problems and systematically analyzes separability across the processing hierarchy. The framework first examines the discriminative potential of the original input data and handcrafted feature representations through a \textit{Foundational Probe}. It then evaluates the quality of the learned latent representations produced by the feature extractor using a \textit{Latent Probe}. Finally, a \textit{Final Probe} assesses the separability achieved in the network's output space, enabling a comprehensive diagnosis of how discriminative information evolves from the input to the final decision layer.

The contributions of this work are threefold. 
First, we propose TriProbe as a stage-consistent diagnostic framework for explainable evaluation of task separability across learning stages. 
Second, we show how it provides actionable insights by diagnosing task difficulty, localizing bottlenecks, and revealing raw-level data complexity that informs both collection strategies and interpretation of results. Third, we demonstrate TriProbe on an sEMG gesture recognition task, where it uncovers bottlenecks consistent with downstream performance and prior studies, while exposing how separability evolves across stages.  

\begin{figure*}[t]
    \centering
    \includegraphics[width=0.7\textwidth]{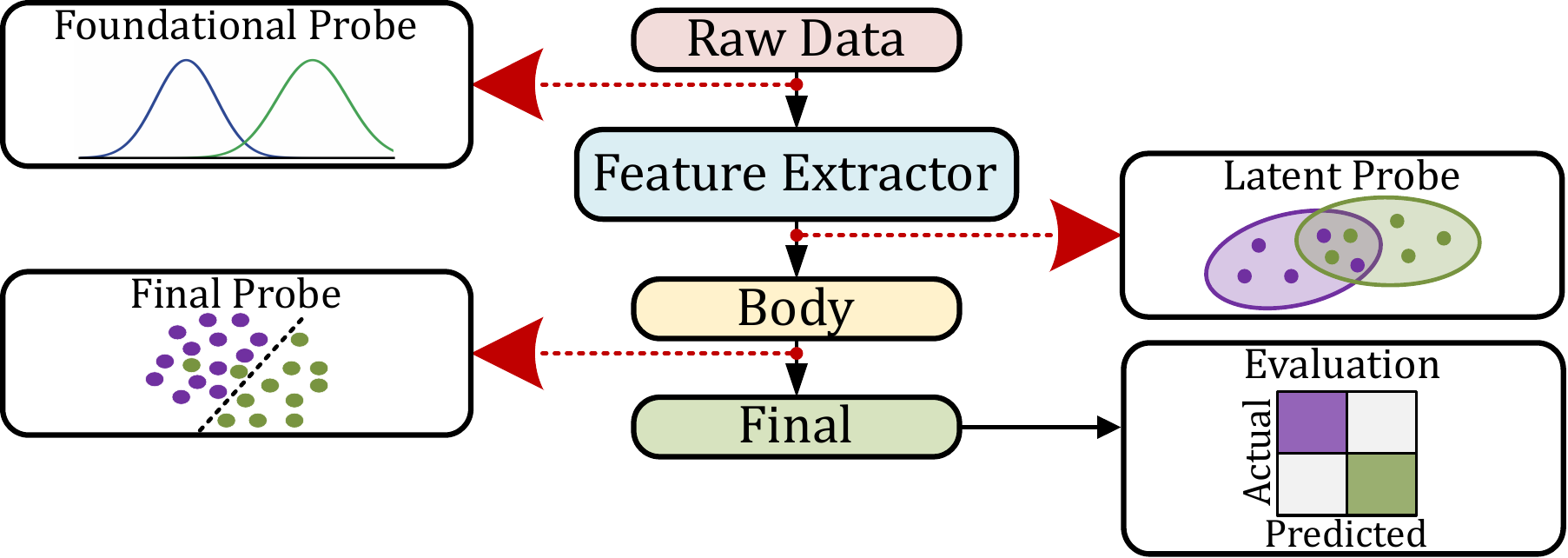}
    \caption{TriProbe: a framework for explainable separability analysis across learning stages.}
    \label{fig:ExAI_framework}
\end{figure*}

\section{Proposed multi-level probing}
\label{sec:proposed}

This work introduces {TriProbe}, a multi-level probing framework for diagnosing the root causes of task separability difficulty across successive stages of learning. The key idea is that separability may degrade or improve at different points in the pipeline, so identifying bottlenecks requires both localizing the stage and pinpointing the most affected class pairs. TriProbe decomposes a multi-task problem into binary sub-tasks, preventing models from exploiting indirect cues from unrelated classes and ensuring that analysis reflects the intrinsic difficulty of each pair. As illustrated in Figure \ref{fig:ExAI_framework}, {TriProbe} employs three complementary probes : (i) a {Foundational Probe} to assess raw data and hand-crafted features, (ii) a {Latent Probe} to evaluate feature extractor representations, and (iii) a {Final Probe} to analyze final-level separability. By combining binary decomposition with multi-level probing, TriProbe offers a principled and interpretable diagnosis of whether limitations arise from data, representation, or final stages.


\subsection{Foundational Probe}
\label{sec:probe1}
Let $\mathbf{X} \in \mathbb{R}^{n \times m}$ denote the raw multi-channel input data, where $n$ is the number of time samples and $m$ the number of input channels. For each class $c_i$, let $\mathbf{X}_{c_i} \in \mathbb{R}^{n_i \times m}$ represent the set of raw samples belonging to that class, with $n_i$ denoting the number of samples. In addition to the raw space, we also consider a hand-crafted feature space. A feature extraction function $\Phi(\cdot)$ maps each raw input $\mathbf{x} \in \mathbb{R}^{n \times m}$ into a feature vector $\mathbf{f} = \Phi(\mathbf{x}) \in \mathbb{R}^{d}$, where $d$ is the number of engineered features (e.g., entropy measures, zero-crossings, RMS, etc.). Collecting across all samples of class $c_i$ yields the feature matrix $\mathbf{F}_{c_i} \in \mathbb{R}^{n_i \times d}$.  


\textbf{Maximum Fisher’s Discriminant Ratio (F1): }
\label{sec:probe1_1}
A central analytical tool in our framework is Maximum Fisher’s Discriminant Ratio (F1), drawn from data complexity analysis~\cite{ho2002complexity}. This metric quantifies separability between two distributions along individual feature dimensions and identifies the single most discriminative feature for distinguishing a target class from a reference. In the one-vs-rest setting, where a target class $c_i$ is compared against all others, the Fisher score of feature $k$ is defined as
\[
\text{F1}_k(c_i) = \frac{(\mu_{i,k} - \mu_{\text{rest},k})^2}{\sigma_{i,k}^2 + \sigma_{\text{rest},k}^2},
\]
and the overall score is $\text{F1}(c_i) = \max_k \text{F1}_k(c_i)$.  
For pairwise analysis between two classes $(c_i,c_j)$, the same formulation applies by replacing the ``rest'' statistics with those of class $c_j$. Higher F1 values indicate stronger separability, while low values suggest intrinsic overlap. This metric is particularly useful because it highlights the single most discriminative feature dimension, allowing us to assess whether poor separability stems from intrinsic data overlap at the raw level.

\subsection{Latent Probe}
\label{sec:probe2}
This probe evaluates task separability in the learned representation space. The feature extractor can be any contemporary architecture, which makes the approach general. In this work, we employ an autoencoder (AE) and use the encoder output after ensuring satisfactory reconstruction quality, while discarding the decoder. This choice avoids bias from downstream task objectives such as classification, while still capturing rich data-driven features. We train a dedicated AE for each class, enabling fine-grained diagnosis and preventing the AE to exploit indirect cues from unrelated classes during reconstruction. This ensures that the encoder learns features solely from its own class distribution, thereby preserving class-specific representation quality. Concretely, the encoder maps each input sample $\mathbf{x}$ into a latent feature vector $\mathbf{f} \in \mathbb{R}^d$. Collecting samples of class $c_i$ forms a feature matrix $\mathbf{F}_{c_i} \in \mathbb{R}^{n_i \times d}$, and similarly $\mathbf{F}_{c_j}$ for class $c_j$, which are then compared to measure pairwise separability in the latent space.

\subsection{Final Probe}
\label{sec:probe3}
This probe is placed immediately before the final separation stage (e.g., the last classifier layer). It aims to evaluate how effectively the body network prepares discriminative representations for the final decision. To maintain generality, we employ a simple multilayer perceptron (MLP) as the classifier body, positioned after AE’s latent representation (encoder output). The MLP is chosen deliberately as a generic architecture, free from domain-specific inductive biases. Therefore, the probe reflects the intrinsic quality of the representations rather than advantages of a specialized classifier. For each input sample, the probe captures the representation vector at the penultimate layer (i.e., the input to the final layer). These vectors form the basis for evaluating task separability at the separation boundary. This enables assessment of whether difficulty arises from insufficiently discriminative representations at the body level, as opposed to intrinsic data limitations (Foundational Probe) or bottlenecks in feature extraction (Latent Probe).

\subsection{Interpretation}
TriProbe provides a stage-wise view of how task separability evolves through the learning pipeline. Low scores at the Foundational Probe signal intrinsic data complexity, while discrepancies between later probes expose architectural limitations. A key guideline is that if separability is strong in the latent space but weak at the final stage, the issue lies in the intervening architecture rather than the data. Thus, TriProbe not only diagnoses task difficulty via binary decomposition and localizes bottlenecks, but also informs data quality, guides architectural refinement, and supports the rational interpretation of results. By pinpointing \emph{where} separability is lost or preserved, TriProbe advances the goals of Explainable AI with a practical tool for both analysis and design.


\section{Experiment}
\label{sec:experiment}

\subsection{Setup}
We evaluate the proposed TriProbe framework on the Roshambo dataset~\cite{roshambo_dataset}, a benchmark known for its low signal-to-noise ratio. The dataset contains recordings from ten participants using a Myo armband with eight sEMG channels sampled at 200~Hz. Participants performed three gestures --- Rock ($R$), Paper ($P$), Scissors ($S$) --- plus a control (Rest). Each participant completed three sessions, with five trials of 3~seconds (s) per gesture, yielding 450 trials in total. To isolate steady-state activity, the first and last 600~ms of each trial were discarded. A sliding window of 400 samples (2\,s) with 50\% overlap was then applied, yielding 296 samples per class of size $8 \!\times\! 400$ (channels $\times$ time).

\textbf{Foundational Probe:} Each $8 \times 400$ sample was transformed into a 72-D feature vector ($8$ channels $\times$ $9$ features) using a modality-agnostic set of hand-crafted features including Shannon entropy, sample entropy, zero crossings, waveform length, root mean square (RMS), slope sign changes, median frequency, wavelet energy, and fractal dimension. This yields $\mathbf{F}_{c_i} \in \mathbb{R}^{296 \times 72}$ per class, capturing temporal, spectral, and complexity characteristics while avoiding domain-specific bias.  
\begin{table*}[ht]
\centering
\caption{The utilized AE. $k$: kernel, $s$: stride, BN: BatchNorm.}
\label{tab:autoencoder}
\renewcommand{\arraystretch}{0.99}
\begin{tabularx}{0.65\textwidth}{|l|X|c|}
\hline
\textbf{Layer} & \textbf{Config} & \textbf{Output} \\
\hline
Input   & -- & $1\times8\times400$ \\
\hline
Conv1   & $1$\hspace{3.00mm}$\!\rightarrow\!128$, $k\!=\!3\!\times\!3$, $s\!=\!1$, \hspace{1.0mm}ReLU,\hspace{1.0mm}\hspace{5.95mm}BN,\hspace{1.0mm}Dropout & $128\times6\times398$ \\
Conv2   & $128\!\rightarrow\!256$, $k\!=\!3\!\times\!3$, $s\!=\!1$, \hspace{1.0mm}Leaky-ReLU,\hspace{1.0mm}BN,\hspace{1.0mm}Dropout & $256\times4\times396$ \\
Conv3   & $256\!\rightarrow\!512$, $k\!=\!3\!\times\!3$, $s\!=\!1$, \hspace{1.0mm}ELU,\hspace{1.0mm}\hspace{6.95mm}BN,\hspace{1.0mm}Dropout & $512\times2\times394$ \\
Conv4   & $512\!\rightarrow\!1$,\hspace{2.75mm}$k\!=\!2\!\times\!3$, $s\!=\!1$ & $1\times1\times392$ \\
\hline
Latent  & -- & $392$ \\
\hline
Deconv1 & $1$\hspace{3.00mm}$\!\rightarrow\!512$, $k\!=\!2\!\times\!3$, $s\!=\!1$, \hspace{1.0mm}ReLU,\hspace{1.0mm}\hspace{5.95mm}BN,\hspace{1.0mm}Dropout & $512\times2\times394$ \\
Deconv2 & $512\!\rightarrow\!256$, $k\!=\!3\!\times\!3$, $s\!=\!1$, \hspace{1.0mm}Softplus,\hspace{1.0mm}\hspace{2.85mm}BN,\hspace{1.0mm}Dropout & $256\times4\times396$ \\
Deconv3 & $256\!\rightarrow\!128$, $k\!=\!3\!\times\!3$, $s\!=\!1$, \hspace{1.0mm}ELU,\hspace{1.0mm}\hspace{6.95mm}BN,\hspace{1.0mm}Dropout & $128\times6\times398$ \\
Deconv4 & $128\!\rightarrow\!1$,$~~~~~~$\hspace{0.2mm}$k\!=\!3\!\times\!3$, $s\!=\!1$ & $1\times8\times400$ \\
\hline
Output  & -- & $1\times8\times400$ \\
\hline
\end{tabularx}
\end{table*}

\textbf{Latent Probe:} Three class-specific AE (\textit{R}-only, \textit{P}-only, \textit{S}-only) were trained to construct latent spaces for pairwise separability analysis. When referring to an AE in the experiments, we specifically use the architecture detailed in Table~\ref{tab:autoencoder}. The encoder output, once achieving satisfactory reconstruction quality, was taken as the latent representation. Each class $c_i$ produces $\mathbf{F}_{c_i} \in \mathbb{R}^{296 \times 392}$, with 392 being the latent dimension. Training used 300 epochs with Huber loss ($\delta=0.25$), Adam optimizer ($\text{lr}=5 \times 10^{-4}$, weight decay $10^{-5}$), and all available samples, aiming the robust representation learning rather than downstream classification.  

\textbf{Final Probe:} For each binary sub-task (\textit{R}–\textit{P}, \textit{R}–\textit{S}, \textit{P}–\textit{S}), we trained a pairwise AE whose encoder output was fed into a simple multilayer perceptron (MLP). To avoid task-objective bias, the AE was kept frozen (decoder discarded, encoder weights fixed), and only the MLP was trained. The initial MLP layers are treated as the body, while the last layer defines the final separator, with the probe placed at their interface. This design preserves architectural neutrality while enabling the probe to identify whether separability limitations arise from insufficient body representations or from the final decision rule.

\subsection{Results}
\begin{figure*}[t]
    \centering
    \includegraphics[width=\linewidth]{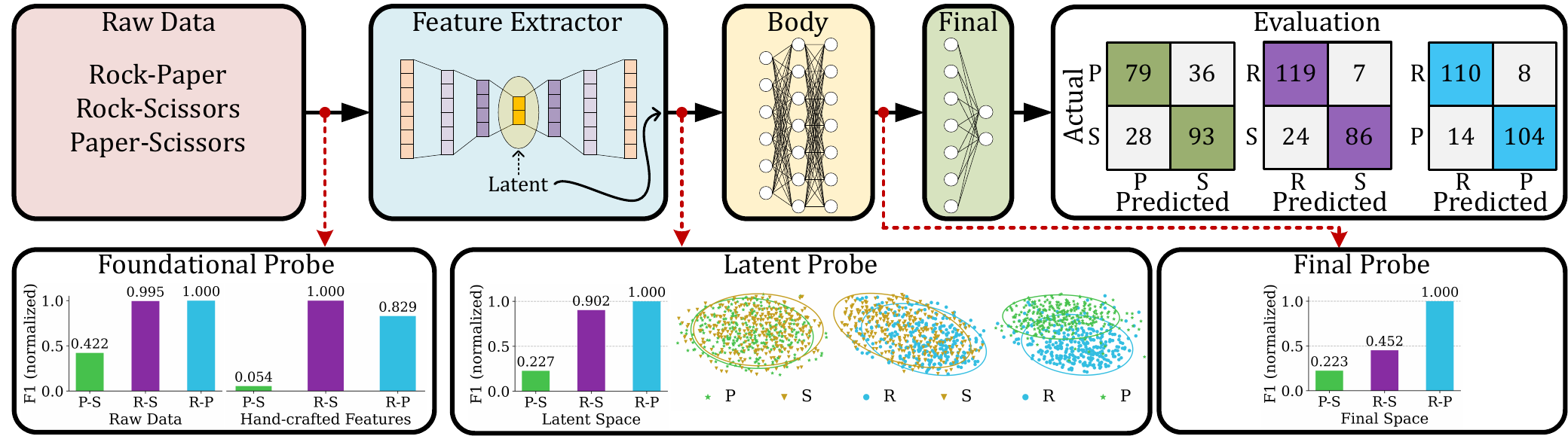}
    \caption{Experimenting the proposed \textbf{TriProbe} framework on the Roshambo dataset, an sEMG-based gesture recognition task with three classes (Rock, Paper, and Scissors). The framework decomposes the problem into binary sub-tasks (Rock–Paper, Rock–Scissors, Paper–Scissors) and evaluates separability across three levels: (i) \textbf{Foundational Probe} measuring raw data and hand-crafted feature discriminability using F1; (ii) \textbf{Latent Probe} assessing latent representations with F1 and visualized using UMAP scatter plots, with covariance ellipses summarizing class spread; and (iii) \textbf{Final Probe} analyzing separability at the penultimate classifier layer. Evaluation confusion matrices are shown for each binary sub-task.}
    \label{fig:triprove_experiment}
\end{figure*}

Figure~\ref{fig:triprove_experiment} illustrates the application of TriProbe on the Roshambo dataset. We first validate the use of F1 for task separability evaluation and show its consistency with both downstream performance and prior results on this dataset. This establishes the analytical core of TriProbe as a robust evaluator, capable of supporting multiple interpretations when deployed at different stages of the learning pipeline. Second, we report  the experiment specific findings.

\begin{figure}[t]
    \centering
    \centering
    \includegraphics[width=0.8\linewidth]{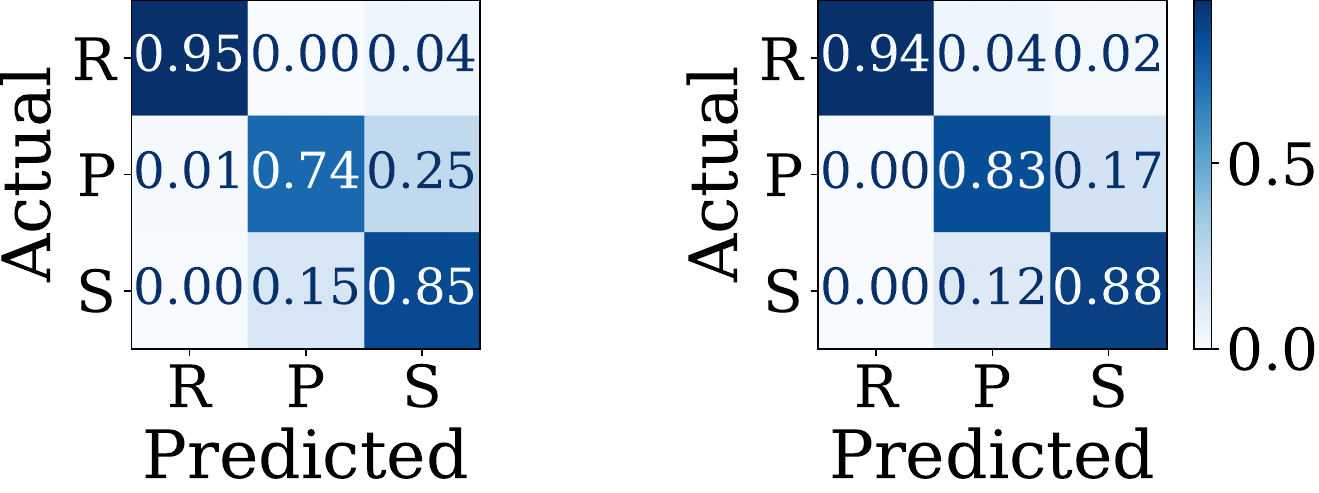}
    \caption{Normalized confusion matrices of two Roshambo model variants from~\cite{garg2020signals}.}
    \label{fig:CM_garg2020signals}
\end{figure}

\textbf{F1 Validation:}
F1 applied at the Foundational probe shows that the \textit{P}–\textit{S} sub-task has the weakest separability, with substantial overlap in both raw data and hand-crafted feature assessments. In contrast, \textit{R}–\textit{S} and \textit{R}–\textit{P} consistently display stronger separability. A closer look reveals a minor discrepancy: raw data suggests \textit{R}–\textit{P} is slightly easier than \textit{R}–\textit{S}, while hand-crafted features suggest the opposite. At the Latent probe, F1 results align with the raw-level findings, and the Final probe largely preserves this pattern, except for a noticeable degradation in \textit{R}–\textit{S}. Confusion matrices further confirm that \textit{P}–\textit{S} is the hardest pair to discriminate, consistent with both our probes and prior results from~\cite{garg2020signals} (see Fig.~\ref{fig:CM_garg2020signals}). Interestingly, their two model variants diverge slightly: one favors \textit{R}–\textit{S}, the other \textit{R}–\textit{P}, echoing our observation that such small gaps are sensitive to the feature space on which the learner operates.

Two key insights emerge: (i) inherent data complexity observed at the raw level propagates through the pipeline, meaning downstream stages should not contradict these trends, and (ii) F1 can flag sub-task separability difficulties early, even at the data collection stage, making it a practical diagnostic tool. Taken together, these results validate F1 as an effective probe of task separability, allowing researchers to peak into the black box.

\textbf{TriProbe Analysis:}
With F1 validated as a reliable separability evaluator, we now leverage it to analyze different stages of the learning pipeline. By decomposing the multi-task problem into binary sub-tasks, TriProbe enables pairwise separability analysis and reveals which sub-tasks act as bottlenecks from raw data through to downstream performance. Our results consistently indicate that separability is most challenging for \textit{P}–\textit{S}, as reflected in both our binary confusion matrices and prior multi-task evaluations in~\cite{garg2020signals}. This shows that TriProbe can diagnose the root cause of multi-task difficulty already at the raw data level. Accordingly, \textit{P}–\textit{S} emerges as the primary bottleneck pair, drawing attention to where learning pipeline design should be strengthened, especially in multi-task settings. 
Beyond bottleneck detection, TriProbe also reveals how separability evolves across stages, supporting architectural search. For instance, \textit{R}–\textit{S} shows good separation in the Latent probe but degrades at the Final probe, suggesting ineffective learning in the MLP body. Redesigning this stage could mitigate the loss. Similarly, \textit{P}–\textit{S} not only appears as the hardest pair overall but also undergoes degradation from Foundational to Latent probe, underscoring the need for stronger feature extraction. Thus, TriProbe not only diagnoses bottlenecks but also tracks the propagation of separability across the pipeline, enabling principled evaluation of whether architectural modifications improve or degrade performance relative to a baseline.

\section{Discussion}
TriProbe provides a structured methodology for diagnosing task separability across the full learning pipeline. Unlike existing probing approaches that focus on isolated representations, TriProbe localizes where separability is gained or lost throughout the entire learning pipeline. While validated here on an sEMG gesture recognition dataset, its formulation is architecture-agnostic and task-independent, suggesting broad applicability across domains such as speech, EEG/MEG, and sensor networks.

While TriProbe builds upon established components such as Fisher's Discriminant Ratio, autoencoders, and probing concepts, its novelty lies in their integration into a unified diagnostic framework for explainable task separability analysis. Rather than introducing a new classifier or separability metric, TriProbe provides a stage-consistent methodology that systematically traces how discriminative information evolves from the raw input space, through learned latent representations, to the final decision space. By combining binary task decomposition with a common separability criterion across all stages, the framework enables direct localization of representational bottlenecks and distinguishes whether limitations originate from the data itself, the feature extractor, or the downstream classifier. This unified perspective is not provided by existing probing or attribution methods, which typically analyze only a single stage or explain individual predictions.

A promising use case for TriProbe is early-stage evaluation during data collection. By probing separability  at the raw and hand-crafted feature level, TriProbe serves as an early warning system, allowing pilot studies and assess data quality before committing to large-scale collection or model training. This is particularly valuable in domains where acquisition is costly. Another application lies in verifying the rationality of experimental findings. If probe results at early stages contradict downstream model performance, TriProbe can highlight potential issues such as overfitting, data leakage, or ineffective architectural choices. In this way, TriProbe complements traditional performance metrics by providing an interpretable explanation of \emph{where} separability is lost or preserved across stages.

Beyond these general applications, our findings also suggest dataset-specific improvements. For example, the observed degradation of the \textit{R}–\textit{S} pair between the latent and final probes indicates that the current MLP body may be ineffective, motivating experiments with alternative architectures. Such targeted refinements illustrate how TriProbe can guide iterative design by localizing weaknesses in the pipeline. Extending this idea, future work should evaluate TriProbe across different datasets and modalities, not only to confirm its generalizability but also to explore how separability evolves under varying data complexities and learning architectures.

\section{Conclusion}
In conclusion, TriProbe contributes to Explainable AI by providing a unified framework for tracing the root causes of task separability difficulty across raw data, learned representations, and decision spaces. Rather than relying solely on downstream performance metrics, it offers stage-wise insights into how discriminative information evolves throughout the learning pipeline, enabling the localization of bottlenecks and the identification of challenging class pairs. Consequently, TriProbe serves not only as an early diagnostic tool for assessing data quality, but also as a means of validating experimental findings and guiding architecture refinement through interpretable separability analysis. 
Although we demonstrated TriProbe here on an sEMG gesture recognition task, the framework is designed to be architecture-agnostic and readily applicable to other machine learning domains. Future work will evaluate TriProbe across additional modalities and datasets, investigate alternative separability measures, and integrate the framework into adaptive learning pipelines, with the long-term goal of establishing a general-purpose, separability-driven methodology for interpretable and reliable AI.

\bibliographystyle{IEEEbib}
\bibliography{References/mohammad, References/refs, References/refs_1}

\end{document}